# CarveMix-RC: Addressing Rare-Class Imbalance Through Lesion-Aware Synthetic Augmentation for Brain Metastasis Segmentation

M. S. Sadique[1] [0000−0002−6734−6802], MD Fayaz Bin Hossen[1] [0009-0008-5109-2466], Michael L. Evans[1] [0009-0002-6880-2950], W. Farzana[1] [0000−0003−1995−2426], Asfaqur Rahman[1] [0009-0007-9624-2766] A. Temtam[1] [0000−0002−1983−4422], and K. M. Iftekharuddin[1,2] [0000−0001−8316−4163]

[1]Vision Lab, Department of Electrical and Computer Engineering, Old Dominion University, Norfolk, VA 23529, USA
[2]Interdisciplinary Schools, Old Dominion University, Norfolk, VA 23529, USA
https://sites.wp.odu.edu/VisionLab/
{msadi002, mhoss006, mevan028, wfarz001, arahm003, atemt001, kiftekha}@odu.edu

**Abstract.** Accurate segmentation of post-treatment brain metastases is essential for treatment planning, longitudinal disease monitoring, and quantitative assessment of therapeutic response. The BraTS-MET 2026 Task 1 challenge introduces a clinically relevant segmentation problem involving four anatomically distinct tumor subregions: non-enhancing tumor core (NETC), surrounding non-enhancing FLAIR hyperintensity (SNFH), enhancing tumor (ET), and the resection cavity (RC). Among these, RC segmentation is particularly challenging because of its low prevalence, heterogeneous postoperative appearance, and lesion-wise evaluation protocol, leading conventional segmentation networks to prioritize dominant tumor classes during optimization.

The proposed nnU-Net-based framework explicitly addresses RC segmentation through four complementary components: (i) RC-weighted Dice and Cross-Entropy optimization to alleviate class imbalance, (ii) anatomically consistent cavity augmentation to increase the diversity of postoperative cavity appearances, (iii) a residual encoder architecture for enhanced multi-scale feature learning, and (iv) lesion-aware morphological post-processing to suppress false-positive cavity predictions while preserving anatomically plausible structures.

The proposed framework was evaluated on the **BraTS-MET 2026 Task 1 online validation benchmark** using multi-parametric MRI. Among the evaluated configurations, the ensemble model (Residual Encoder nnU-Net + nnU-Net + RC-aware CarveMix) achieved the best performance, with lesion-wise Dice scores of **0.732, 0.752, 0.708, and 0.575** and corresponding NSD scores of **0.794, 0.798, 0.727, and 0.474** for ET, TC, WT, and RC, respectively. These experimental results show that integrating RC-aware optimization, anatomically consistent augmentation, and lesion-aware post-processing provides an effective strategy for improving rare resection cavity segmentation in post-treatment brain metastases. Code and implementation details are publicly available at https://github.com/shiblyg/carvemix-rc.

# 1 Introduction

Brain metastases (BMs) are the most common intracranial malignancies in adults, affecting approximately 20-40% of patients with systemic cancer and representing a major cause of neurological morbidity and mortality [1-3]. Advances in surgery, stereotactic radiosurgery (SRS), immunotherapy, and targeted therapies have improved patient survival, increasing the need for longitudinal MRI to monitor treatment response and disease progression. Accurate segmentation of post-treatment tumor subregions is therefore essential for treatment planning, response assessment, recurrence monitoring, and quantitative imaging biomarker development.

Post-treatment brain metastasis segmentation remains considerably more challenging than segmentation of untreated tumors because surgical intervention fundamentally alters normal anatomy. Postoperative cavities, blood products, edema, tissue deformation, and treatment-related imaging changes introduce substantial anatomical and appearance variability that can closely resemble recurrent disease or radiation-induced effects, making reliable delineation difficult even for experienced neuroradiologists [4]. These challenges motivate the BraTS-MET 2026 Challenge, which provides a standardized benchmark for post-treatment brain metastasis segmentation using multi-parametric MRI.

BraTS-MET Task 1 requires simultaneous segmentation of four clinically relevant tissue classes: the non-enhancing tumor core (NETC), surrounding non-enhancing FLAIR hyperintensity (SNFH), enhancing tumor (ET), and resection cavity (RC). Among these, RC is particularly challenging because it occupies a relatively small image volume while exhibiting substantial variability in morphology and appearance due to differences in surgical technique, healing stage, hemorrhage, cerebrospinal fluid accumulation, and surrounding tissue response. As a result, RC is severely underrepresented during optimization, often leading to reduced sensitivity and inconsistent lesion detection in conventional segmentation models [5-9].

Recent advances in medical image segmentation have been largely driven by nnU-Net, which automatically configures preprocessing, network architecture, and training strategies for robust performance across diverse biomedical imaging tasks. However, its standard optimization and sampling strategy does not explicitly address severe foreground class imbalance. Consequently, optimization is dominated by abundant tumor regions, encouraging the network to focus on ET, TC, and WT while underrepresenting the rare RC class. This limitation becomes particularly evident in lesion-wise evaluation, where even a small missing resection cavity can substantially reduce overall performance despite accurate segmentation of larger tumor regions [10,11].

Addressing postoperative brain metastasis segmentation requires more than increasing network capacity. The rarity of resection cavities introduces a fundamental learning challenge: models are inherently biased toward anatomically dominant tumor regions, resulting in poor representation of uncommon postoperative structures. Although data augmentation, class-balanced optimization, and post-processing have each been explored to mitigate class imbalance, they are typically investigated in isolation and do not explicitly account for the anatomical characteristics of postoperative cavities. Consequently, accurate segmentation of resection cavities remains a persistent limitation in existing methods.

To address these challenges, we propose a unified lesion-aware segmentation framework that improves the representation and delineation of rare postoperative anatomical structures while preserving robust performance across all tumor subregions. Rather than relying on a single strategy, the framework integrates complementary mechanisms for class-aware optimization, anatomically consistent cavity augmentation, enhanced feature representation, and lesion-aware false-positive suppression within the nnU-Net paradigm. Together, these components improve learning of the underrepresented resection cavities while maintaining anatomical consistency throughout the segmentation process.

The proposed framework was evaluated on the BraTS-MET 2026 Task 1 benchmark using multi-parametric MRI. Experimental results demonstrate consistent improvements in resection cavity segmentation while maintaining competitive performance for enhancing tumor, tumor core, and whole tumor segmentation, indicating that integrating lesion-aware augmentation with class-aware optimization provides an effective solution for postoperative brain metastasis segmentation.

Our main contributions are summarized as follows:

1. Identification of resection cavity segmentation as the main challenge in post-treatment brain metastasis analysis and present a unified framework to address this.
2. Introducing an RC-weighted strategy with anatomically consistent cavity augmentation to enhance learning of underrepresented cavity regions while maintaining performance on dominant tumor sub-regions.
3. Post-processing incorporating residual encoder-based feature learning alongside lesion-aware morphological post-processing to enhance segmentation robustness and minimize false-positive cavity predictions.

# 2 Related Work

## 2.1 Brain Metastasis Segmentation

Brain metastases (BMs) represent the most prevalent intracranial malignancy in adults and exhibit considerable variability in size, morphology, anatomical location, and treatment response. Accurate delineation of metastatic lesions and postoperative tissue compartments is essential for stereotactic radiosurgery planning, longitudinal disease monitoring, and quantitative response assessment. While considerable progress has been achieved in automated brain tumor segmentation using deep learning, most existing studies [5,7] have focused on untreated gliomas or preoperative metastatic lesions, where tumor boundaries are comparatively well defined. In contrast, post-treatment brain metastases introduce substantially greater anatomical complexity due to surgical resection, postoperative cavity formation, hemorrhage, tissue deformation, edema, and heterogeneous enhancement patterns, making accurate segmentation considerably more challenging [5].

The Brain Tumor Segmentation (BraTS) challenges [5] have significantly advanced the development of automated segmentation methods by providing standardized datasets, evaluation protocols, and benchmark leaderboards. While previous BraTS editions primarily emphasized glioma segmentation, the recently introduced BraTS-MET benchmark extends these efforts to pre and post-treatment brain metastases, requiring simultaneous delineation of the non-enhancing tumor core (NETC), the surrounding non-enhancing FLAIR hyperintensity (SNFH), the enhancing tumor (ET), and the resection cavity (RC). This benchmark highlights the unique challenges associated with postoperative anatomy and lesion-wise evaluation [5,7].

## 2.2 Deep Learning for Medical Image Segmentation

Convolutional neural networks have become the dominant paradigm for medical image segmentation, with encoder-decoder architectures such as U-Net establishing the foundation for numerous subsequent developments. Variants incorporating residual learning, attention mechanisms, dense connectivity, transformer-based feature extraction, and multi-scale context aggregation have consistently improved segmentation performance across diverse imaging modalities [12-14]. Despite these architectural advances, segmentation accuracy remains strongly influenced by optimization strategy, training data diversity, and class distribution, particularly in highly imbalanced medical imaging problems [15-18].

Among contemporary segmentation frameworks, **nnU-Net** has emerged as the de facto baseline because of its automatic configuration of preprocessing, architecture selection, patch size, normalization, data augmentation, and training schedules. Rather than proposing a new network architecture, nnU-Net demonstrates that carefully optimized training protocols often outperform increasingly complex architectural modifications. Consequently, nnU-Net has become the reference framework for numerous international segmentation challenges, including multiple BraTS competitions [10,11].

However, the default nnU-Net optimization strategy assumes a relatively balanced set of foreground classes. Under severe class imbalance, optimization gradients become dominated by frequent anatomical structures, leading to reduced sensitivity for minority classes. This limitation is particularly evident for postoperative resection cavities, which occupy only a small fraction of the imaging volume yet contribute substantially to lesion-wise evaluation metrics.

### 2.3 Learning Under Severe Class Imbalance

Class imbalance remains a major challenge in medical image segmentation because lesion classes often differ substantially in size and frequency. Existing approaches address this problem through weighted loss functions, adaptive sampling, hard example mining, and curriculum learning [12,14,19]. While these strategies improve optimization, they cannot compensate for the limited anatomical diversity of rare postoperative structures. Consequently, improving optimization alone is often insufficient for reliable segmentation of underrepresented classes such as resection cavities.

### 2.4 Data Augmentation for Rare Anatomical Structures

Data augmentation plays a central role in improving the robustness and generalization of deep segmentation networks. Conventional augmentation techniques-including random rotations, scaling, elastic deformations, intensity perturbations, and spatial cropping-primarily increase appearance diversity without fundamentally altering anatomical composition. Recently, region-level augmentation strategies such as CutMix, MixUp, ClassMix, and CarveMix have demonstrated improved learning by exchanging semantically meaningful image regions between training samples [20-23]. While these approaches increase foreground diversity, they are not specifically designed for post-operative anatomy. Resection cavities exhibit irregular geometry, heterogeneous signal characteristics, and complex interactions with surrounding edema and residual tumor tissue. Therefore, augmentation strategies for post-operative segmentation should preserve anatomical plausibility while increasing representation of rare cavity appearances. Therefore, we leverage region-based cavity augmentation within a post-operative brain metastasis segmentation framework to improve representation of the underrepresented RC class during training.

### 2.5 Lesion-Aware Segmentation and Postprocessing

Medical image segmentation is now often assessed using lesion-wise metrics that focus on accurately detecting individual pathological structures, rather than just voxel-wise overlap. In these evaluation methods, false-positive predictions and missed small lesions can significantly reduce overall performance, even when global Dice scores appear satisfactory. Consequently, morphology-aware postprocessing remains an important component of modern segmentation pipelines.

Connected-component analysis, size-based filtering, topology-preserving pos-processing, and morphological operations have been widely employed to suppress spurious predictions while preserving anatomically plausible structures. Several BraTS-winning approaches [10,11,13] have demonstrated that simple, yet carefully designed postprocessing strategies can produce measurable improvements, particularly for small or infrequent lesion classes. post-processing. Motivated by these observations, the proposed framework integrates lesion-aware post-processing as its final phase to enhance the anatomical consistency of resection cavity predictions while preserving segmentation accuracy in the remaining tumor regions.

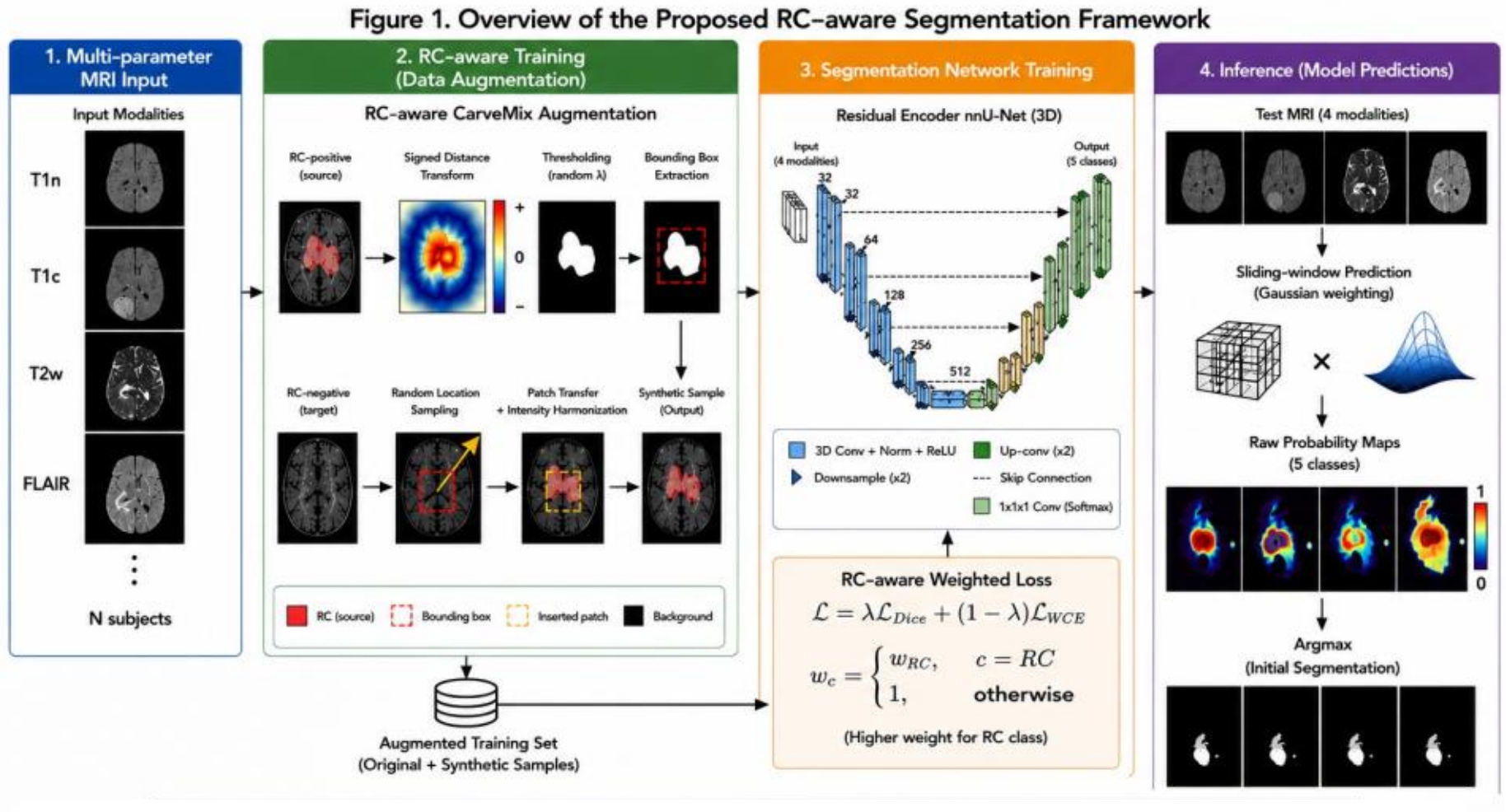

**Fig. 1. Overview of the proposed RC-aware segmentation framework**

### 2.6 CarveMix Augmentation

Postoperative resection cavities are underrepresented in the training data and exhibit substantial variability in size, shape, and anatomical location. To increase the diversity of cavity appearances, we incorporated CarveMix [20], an anatomy-aware augmentation strategy. As illustrated in Figure 1, CarveMix generates additional training samples by transplanting resection cavity regions between anatomically compatible subjects while preserving the corresponding segmentation labels. Unlike conventional intensity or geometric augmentations, CarveMix directly augments the morphology and spatial distribution of postoperative cavities, increasing the variability of rare RC examples presented during training. The augmented samples are used together with the original training data during nnU-Net optimization.

## 3 Methodology

**Fig. 1** illustrates the overall framework of the proposed method. Starting from multiparametric MRI volumes, we first increase the representation of the underrepresented resection cavity (RC) class using an anatomically consistent cavity synthesis strategy adapted from CarveMix. The augmented dataset is subsequently used to train a Residual Encoder nnU-Net with RC-aware optimization. During inference, lesion-aware morphological post-processing is applied to suppress anatomically implausible false-positive cavity predictions while preserving tumor structures.

### 3.1 RC-aware Optimization

The BraTS-MET dataset exhibits severe class imbalance, with the resection cavity accounting for only a small proportion of the foreground voxels. Consequently, standard optimization tends to prioritize larger anatomical structures, such as the enhancing tumor (ET), the non-enhancing tumor core (NETC), and the surrounding non-enhancing FLAIR hyperintensity (SNFH), resulting in inferior RC segmentation.

Given the four MRI modalities

$$X = \{T1, T1c, T2w, T2f\},$$

The segmentation network predicts voxel-wise posterior probabilities

$$P = f_\theta(X).$$

The network is optimized using a weighted hybrid objective

$$\mathcal{L} = \lambda\mathcal{L}_{Dice} + (1 - \lambda)\mathcal{L}_{WCE},$$

where the weighted cross-entropy is defined as

$$\mathcal{L}_{WCE} = -\sum_{c=1}^{C} w_c\, g_c \log(p_c),$$

with an increased class weight assigned to the RC category,

$$w_c = \begin{cases} w_{RC}, & c = RC, \\ 1, & \text{otherwise.} \end{cases}$$

This weighting increases the contribution of rare cavity voxels to optimization while maintaining stable learning in the remaining tumor compartments. In addition, nnU-Net foreground patch oversampling is increased to expose the network to RC-containing regions more frequently during training. In all RC-aware experiments, Dice and cross-entropy were equally weighted ($w_{Dice}$= $w_{CE}$ = 1), the cross-entropy class weights were [1, 1, 1, 1, 3] ($w_{RC}$= 3), and the foreground oversampling was increased from 0.33 to 0.66.

### 3.2 RC-aware Anatomically Consistent Cavity Synthesis

Although weighted optimization enhances gradient allocation, the limited number of RC-positive subjects constrains the diversity of cavity appearances encountered during training. To alleviate this limitation, we leverage CarveMix for postoperative cavity synthesis. For the primary augmentation setting, 300 synthetic RC-positive cases were generated; an extended setting used 450 synthetic cases. Donor cases with fewer than 50 RC voxels were excluded, and the signed-distance threshold parameter was sampled as lambda $\sim U(-3, 5)$, where negative and positive values contract and expand the carved RC region, respectively.

For an RC-positive source image, the cavity mask is converted into a signed distance representation

$$D(\boldsymbol{R}_s),$$

and a random threshold

$$\lambda \sim \frac{1}{2}U(\lambda_l, 0) + \frac{1}{2}U(0, \lambda_u)$$

is sampled to generate a variable cavity region

$$M(v) = \begin{cases} \mathbf{1}, & (\boldsymbol{D}(\boldsymbol{R_s})(\boldsymbol{v}) \leq \boldsymbol{\lambda}) \\ \mathbf{0}, & \boldsymbol{otherwise} \end{cases}$$

Unlike the original CarveMix formulation, CarveMix-RC does not rely on source-target spatial correspondence. The carved RC is transferred at its native dimensions using its minimum bounding box to a uniformly sampled location for which the complete box lies within the target volume; infeasible placements are rejected, with no resizing or interpolation. Only the RC mask and corresponding four-modality intensities are transferred, and RC takes precedence over existing labels within the insertion mask; no atlas-based placement or tumor-overlap exclusion is imposed. Before insertion, source intensities are harmonized independently for each modality by matching the local mean and standard deviation of the target region. The resulting synthetic NIfTI cases subsequently undergo standard nnU-Net v2 preprocessing, with the complete procedure defined in Algorithm 1.

**Algorithm 1: RC-aware CarveMix Augmentation**

**Input: Training dataset $\mathcal{D} = \{(X_i, Y_i)\}_{i=1}^{N}$, RC label $c_{RC}$, desired number $T$of synthetic cases.**
**Output: Synthetic dataset $\mathcal{D}_s = \{(\widetilde{X}, \widetilde{Y})\}$.**

1 for $t = 1,2,\ldots,T$do
2 Randomly select an RC-positive source $(X_s, Y_s)$ containing at least 50 RC voxels and an RC-negative target $(X_t, Y_t)$.
3 Define the binary RC mask

$$R_s(v) = \mathbb{1}[Y_s(v) = c_{RC}].$$

4 Compute the signed distance transform $D(R_s)$, negative inside the RC and positive outside.
5 Sample the cavity-size parameter

$$\lambda \sim \mathcal{U}(-3, 5).$$

6 Construct the variable carved region

$$M_s(v) = \mathbb{1}[D(R_s)(v) < \lambda].$$

7 Compute the minimum axis-aligned bounding box $B_s$ enclosing $M_s$.
8 Crop the multimodal source patch $P_X = X_s[B_s]$ and binary mask $P_M = M_s[B_s]$.
9 **if $B_s$ exceeds the target dimensions along any axis then**
10 reject the pair and resample; no resizing or interpolation is performed.
11 **end if**
12 Sample a valid insertion origin q such that the complete patch lies within $X_t$, and let $B_t(q)$ denote the corresponding target box.
13 For each modality m, harmonize source voxels inside $P_M$to the local target statistics:

$$P'_{X,m} = \frac{P_{X,m} - \mu_{s,m}}{\sigma_{s,m} + \varepsilon}(\sigma_{t,m} + \varepsilon) + \mu_{t,m}, \quad \varepsilon = 10^{-6}.$$

The source and target statistics are evaluated only over voxels selected by $P_M$.
14 Initialize $\tilde{X} \leftarrow X_t$ and, within $B_t(q)$, update

$$\tilde{X}_m[B_t(q)] = P'_{X,m} \odot P_M + X_{t,m}[B_t(q)] \odot (1 - P_M).$$

15 Initialize $\tilde{Y} \leftarrow Y_t$ and assign

$$\tilde{Y}[B_t(q)](v) = c_{RC} \quad for\ all\ v\ such\ that\ P_M(v) = 1,$$

while all remaining target labels are unchanged.
16 Add $(\tilde{X}, \tilde{Y})$ to $\mathcal{D}_s$.
17 **end for**
18 **return $\mathcal{D}_s$.**

### 3.3 Residual Encoder Segmentation and Lesion-aware Post-processing

The augmented dataset is used to train a Residual Encoder nnU-Net while preserving the automated configuration and training pipeline of nnU-Net. During inference, sliding-window prediction with Gaussian weighting generates voxel-wise probability maps, which are subsequently converted to discrete segmentation labels. Lesion-aware post-processing applies 26-connected component analysis: RC components <50 voxels and ET components <15 voxels are removed. Binary hole filling is then applied to the RC segmentation to reduce internal discontinuities. Together, these operations suppress small isolated false-positive predictions while preserving the boundaries of larger predicted lesions.

## 4 Experimental Setup

### 4.1 Dataset

Experiments were conducted on the **BraTS-MET 2026 Task 1** dataset, which comprises 1,296 pre- and post-treatment brain metastasis cases across four co-registered MRI modalities (T1, T1c, T2w, and T2f). Expert annotations include **four semantic labels**: non-enhancing tumor core (NETC), surrounding non-enhancing FLAIR hyperintensity (SNFH), enhancing tumor (ET), and resection cavity (RC).

### 4.2 Implementation Details

The proposed framework was implemented using **nnU-Net v2** with a Residual Encoder backbone in **PyTorch**. Images were preprocessed using the default nnU-Net pipeline, including foreground cropping, intensity normalization, and automatic target spacing estimation. Training employed weighted Dice and Cross-Entropy loss with increased emphasis on the RC class, RC-aware patch oversampling, and anatomically consistent cavity augmentation. Unless otherwise specified, all remaining hyperparameters followed the default nnU-Net configuration. Training and inference were performed on NVIDIA H100 GPUs.

Inference was performed using nnU-Net's sliding-window prediction with Gaussian weighting and test-time augmentation. Final predictions were refined using lesion-aware post-processing (**RC-Weighted nnU-Net)**, including connected-component filtering, cavity hole filling, and morphology-based removal of anatomically implausible RC predictions. Model ensembling (**RC-Weighted ResEncL Predictions)** was performed by averaging the softmax probabilities, followed by lesion-aware post-processing.

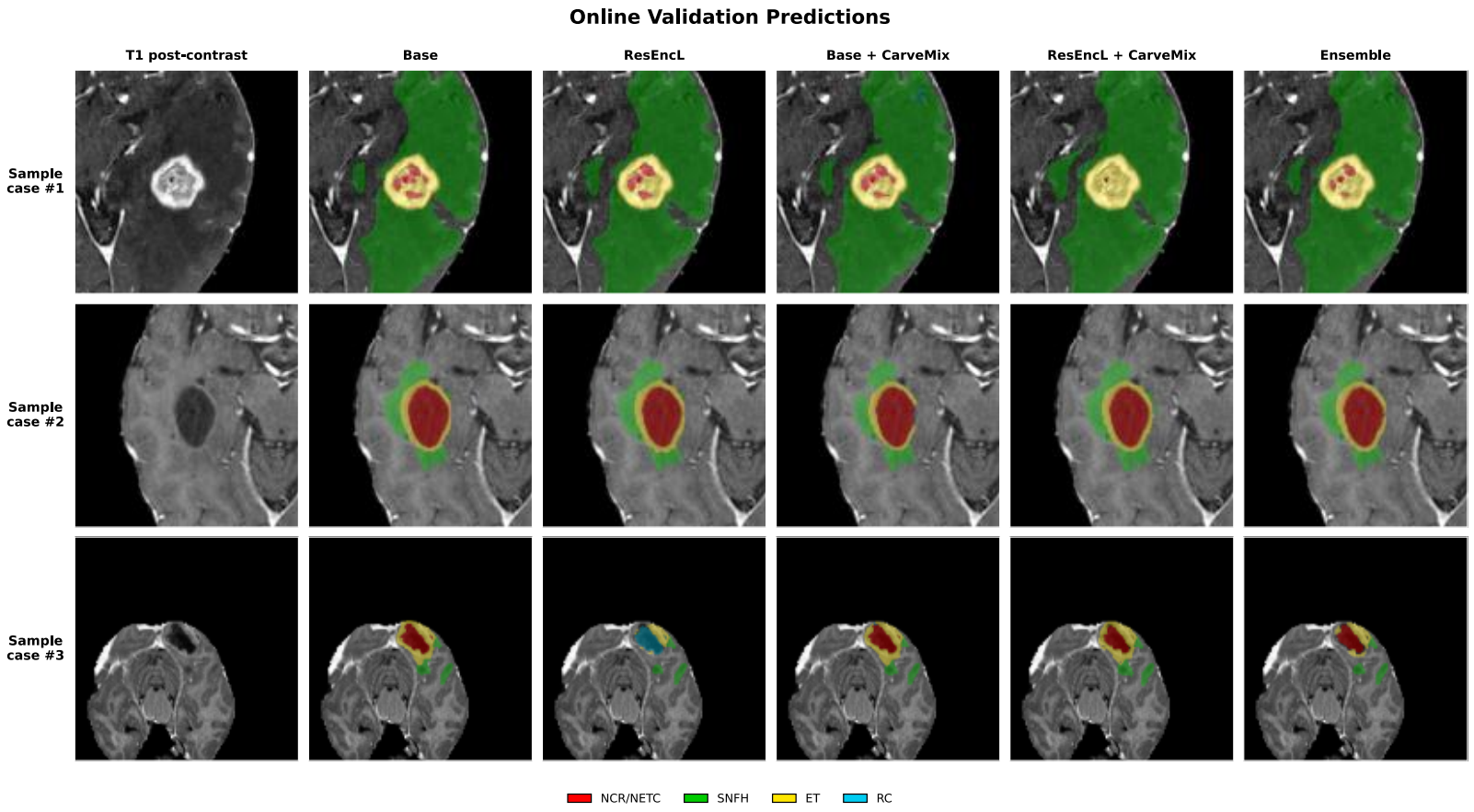

**Fig. 2.** Qualitative Comparison of Five Model Configurations on the BraTS-MET 2026 Online Validation Set

# 5 Results

## 5.1 Experimental Evaluation

The proposed framework was evaluated on the **BraTS-MET 2026 Task 1** validation dataset using the official challenge evaluation server. Performance was evaluated using the official **BraTS-MET 2026** evaluation framework. Segmentation accuracy was assessed using lesion-wise **Dice Similarity Coefficient (DSC)** and **Normalized Surface Dice (NSD)** for ET, RC, TC, and WT. Lesion-detection performance was further evaluated using lesion-wise true positives (TP), false positives (FP), false negatives (FN), and F1-Scores for all, large, and small lesions [24,25]. Since the postoperative resection cavity is the primary focus of this work, particular emphasis is placed on RC segmentation performance while monitoring its effects on the remaining tumor subregions.

## 5.2 Quantitative Results

Table 1 summarizes nine configurations evaluating RC-weighted optimization, CarveMix-RC augmentation, lesion-aware post-processing, and ensembling. On the ResEnc backbone, RC weighting increased RC Dice/NSD from 0.404/0.341 (C2) to 0.479/0.418 (C3), with RC Dice reaching 0.540 after post-processing (C4). CarveMix-RC further increased RC Dice from 0.540 (C4) to 0.575 (C8), although ET, TC, and WT Dice decreased modestly by 0.017, 0.021, and 0.008, respectively; RC NSD remained similar (0.449 vs. 0.450). The benefit of augmentation was therefore concentrated primarily in RC overlap, with a modest trade-off in the more prevalent tumor regions.

Post-processing also improved cavity prediction, increasing RC Dice from 0.530 (C6) to 0.575 (C8) in the ResEnc CarveMix configuration. An equal-size ensemble without CarveMix does not improve upon the corresponding CarveMix ensemble, providing additional control for the effect of model ensembling; detailed results are reported in the Supplementary Material. The final ensemble (C9) achieved Dice/NSD of 0.732/0.794 for ET, 0.752/0.798 for TC, 0.708/0.727 for WT, and 0.575/0.474 for RC.

Table 2 summarizes Lesion-level performance, which showed a pronounced dependence on lesion size. Recall for large ET, TC, and WT lesions was 0.881, 0.883, and 0.858, respectively, compared with 0.194, 0.198, and 0.159 for small lesions. No small RC lesions were detected, whereas large-RC recall reached 0.833, identifying small-lesion sensitivity as the principal remaining limitation. Detailed TP, FP, FN, precision, recall, and F1 analyses are provided in the Supplementary Material.

Paired patient-clustered bootstrap analysis with 10,000 resamples was used to quantify uncertainty. Relative to C2, C9 increased lesion-wise Dice by +0.055 for ET (95% CI, +0.035 to +0.078), +0.053 for TC (+0.033 to +0.076), +0.064 for WT (+0.039 to +0.092), and +0.171 for RC (+0.067 to +0.285). Component-level comparisons and region-specific uncertainty estimates are reported in the Supplementary Material.

**Table 1.** Quantitative comparison of representative model configurations on the BraTS-MET 2026 validation dataset across tumor sub-regions: enhanced (ET), resection cavity (RC), tumor core (TC), whole tumor (WT)

| **Configuration** | **Model** | **ET** | | **TC** | | **WT** | | **RC** | |
|---|---|---|---|---|---|---|---|---|---|
| | | **Dice** | **NSD** | **Dice** | **NSD** | **Dice** | **NSD** | **Dice** | **NSD** |
| Baseline | **Std** | 0.665 | 0.729 | 0.688 | 0.738 | 0.646 | 0.67 | 0.368 | 0.247 |
| Baseline | **Res** | 0.677 | 0.738 | 0.699 | 0.747 | 0.643 | 0.666 | 0.404 | 0.341 |
| + RC weighting | **Res** | 0.662 | 0.725 | 0.686 | 0.736 | 0.647 | 0.667 | 0.479 | 0.418 |
| + RC weighting + PP | **ResEncL** | 0.728 | 0.789 | 0.748 | 0.796 | 0.693 | 0.711 | 0.54 | 0.449 |
| + RC-w + Carve-Mix | **Std** | 0.657 | 0.717 | 0.676 | 0.723 | 0.645 | 0.668 | 0.434 | 0.355 |
| + RC-w + Carve-Mix | **Res** | 0.715 | 0.779 | 0.736 | 0.786 | 0.688 | 0.709 | 0.53 | 0.427 |
| + RC-w + Carve-Mix + PP | **Std** | 0.677 | 0.743 | 0.705 | 0.758 | 0.67 | 0.694 | 0.481 | 0.4 |
| + RC-w + Carve-Mix + PP | **Res** | 0.711 | 0.773 | 0.727 | 0.774 | 0.685 | 0.707 | **0.575** | 0.45 |
| **Ensemble (Std + Res)** | **Both** | **0.732** | **0.794** | **0.752** | **0.798** | **0.708** | **0.727** | **0.575** | **0.474** |

**Table 2.** Lesion-level detection performance of the final ensemble stratified by lesion size. TP/FP/FN and F1 are from the official BraTS-METS evaluation; precision and recall are derived from the reported aggregate counts.

| Region | Lesion size | TP | FP | FN | Preci-sion | Recall | F1 |
|---|---|---|---|---|---|---|---|
| ET | Small | 1.31 | 0.57 | 5.41 | 0.695 | 0.194 | 0.347 |
| | Large | 3.57 | 0.35 | 0.48 | 0.91 | 0.881 | 0.781 |
| | **All** | **4.13** | **0.35** | **2.74** | **0.921** | **0.601** | **0.811** |
| TC | Small | 1.32 | 0.53 | 5.37 | 0.715 | 0.198 | 0.354 |
| | Large | 3.58 | 0.33 | 0.47 | 0.916 | 0.883 | 0.784 |
| | **All** | **4.14** | **0.33** | **2.68** | **0.926** | **0.607** | **0.818** |
| WT | Small | 0.96 | 0.79 | 5.07 | 0.549 | 0.159 | 0.251 |
| | Large | 3.41 | 0.47 | 0.56 | 0.879 | 0.858 | 0.793 |
| | **All** | **3.87** | **0.47** | **2.66** | **0.892** | **0.593** | **0.805** |
| RC | Small | 0 | 0 | 1.17 | -- | 0 | 0 |
| | Large | 0.11 | 0.14 | 0.02 | 0.444 | 0.833 | 0.1 |
| | **All** | **0.12** | **0.14** | **0.06** | **0.457** | **0.677** | **0.476** |

### 5.3 Discussion

The experimental results demonstrate that RC remains the most challenging region in post-treatment brain metastasis segmentation, reflecting its limited representation and substantial variability in postoperative appearance. The proposed combination of RC-weighted optimization, CarveMix-RC augmentation, and lesion-aware post-processing specifically addresses this imbalance during training and inference. CarveMix-RC primarily improved RC segmentation, although modest changes in ET, TC, and WT indicate a trade-off when increasing emphasis on the rare cavity class. The ensemble mitigated this trade-off and provided a more balanced performance across the four target regions while preserving the improvement in RC.

Lesion-level analysis further showed that performance was strongly dependent on lesion size. Larger lesions were detected reliably, whereas small lesions, particularly small RCs, remained the dominant source of error. This suggests that increasing the representation of RC examples alone is insufficient to resolve the small-lesion problem. In addition, CarveMix-RC uses geometrically feasible cavity placement without explicit spatial or atlas-based constraints, which may limit the anatomical realism of some synthetic examples. Future work will investigate spatially informed cavity augmentation and sampling strategies that better represent small postoperative lesions.

**Acknowledgments.** The authors would like to acknowledge partial support for this work by the National Institute of Health grant #R01 EB020683.